\documentclass[runningheads]{llncs}
\usepackage{graphicx}
\usepackage{subfigure}
\usepackage{import}
\usepackage{setspace}
\usepackage{geometry} % added 27-02-2014 Markus Englich
\usepackage{epstopdf}
\usepackage[labelsep=period]{caption}  % added 14-04-2016 Markus Englich - Recommendation by Sebastian Brocks

\usepackage{hyperref}
\usepackage{graphicx}
\usepackage[export]{adjustbox}
\usepackage{amsmath,amsfonts,amssymb}
\usepackage[acronyms]{glossaries}
\usepackage{color, soul}
\usepackage{diagbox}	% diagonal box
\usepackage{enumitem}
\usepackage{gensymb}
\usepackage[nameinlink, capitalize]{cleveref}
\usepackage{hhline}
\usepackage{inputenc}
\usepackage{todonotes}
\usepackage{algorithm}
\usepackage[noend]{algpseudocode}
\usepackage{float}
\usepackage{cuted} %main package: to insert the image 
\usepackage[percent]{overpic}
\usepackage{tabularx} % in the preamble
\usepackage{multirow}

\newcommand\blfootnote[1]{%
  \begingroup
  \renewcommand\thefootnote{}\footnote{#1}%
  \addtocounter{footnote}{-1}%
  \endgroup
}

\usepackage{xspace}

\newcommand{\approxi}{approx.\ }

\newcommand{\m}{\,m\xspace}

\newcommand{\s}{\,s\xspace}

\newcommand{\fps}{\,fps\xspace}

\newcommand{\percent}{\,\%\xspace}

\newacronym{ASG}{ASG}{{Average Shading Gradient}}

\newacronym[shortplural={CNNs}, firstplural={convolutional neural networks (CNNs)}, longplural={convolutional neural networks}]{CNN}{CNN}{convolutional neural network}
\newacronym{COTS}{COTS}{commercial off-the-shelf}
\newacronym[shortplural={CRFs}, longplural={{Conditional Random Fields}}]{CRF}{CRF}{{Conditional Random Field}}
\newacronym{CT}{CT}{census transform}

\newacronym{DLT}{DLT}{{Direct Linear Transformation}}
\newacronym{DoG}{DoG}{{Difference of Gaussian}}

\newacronym{EPnP}{EPnP}{{Efficient Perspective-n-Point}}

\newacronym{GPGPU}{GPGPU}{general purpose computation on a {GPU}}
\newacronym{GPS}{GPS}{{Global Positioning System}}
\newacronym{GTA}{GTA V}{Grand Theft Auto V}

\newacronym{ICP}{ICP}{{Iterative-Closest-Point}}
\newacronym{IMU}{IMU}{{Inertial Measurement Unit}}
\newacronym{INS}{INS}{{Inertial Navigation System}}

\newacronym{LIDAR}{LiDAR}{{Light Detection and Ranging}}
\newacronym{L1-rel}{$\text{L1-rel}$}{{relative $\text{L1}$-Norm}}
\newacronym{L1-abs}{$\text{L1-abs}$}{{absolute $\text{L1}$-Norm}}

\newacronym[shortplural={MRFs}, longplural={{Markov Random Fields}}]{MRF}{MRF}{{Markov Random Field}}

\newacronym{MVS}{MVS}{Multi-View-Stereo}

\newacronym{NCC}{NCC}{normalized cross correlation}

\newacronym{PCL}{PCL}{{Point Cloud Library}}

\newacronym{RANSAC}{RANSAC}{{Random Sampling Consensus}}
\newacronym{RMSE}{RMSE}{{Root Mean Square Error}}
\newacronym{MAE}{MAE}{{Mean Absolute Error}}

\newacronym[shortplural={ROIs}, longplural={regions of interest}]{ROI}{RoI}{region of interest}

\newacronym{SAD}{SAD}{sum of absolute differences}
\newacronym{SFM}{SfM}{{Structure-from-Motion}}
\newacronym{SGBM}{SGBM}{{Semi-Global Block Matching}}
\newacronym{SGM}{SGM}{{Semi-Global Matching}}
\newacronym{SLAM}{SLAM}{simultaneous localization and mapping}
\newacronym{SMDE}{SMDE}{{Self-supervised Monocular Depth Estimation}}
\newacronym{SSIM}{SSIM}{Structural Similarity}
\newacronym[shortplural={STNs}, longplural={Spatial Transformer Networks}]{STN}{STN}{{Spatial Transformer Network}}
\newacronym[shortplural={surfels}, longplural={surface-elements}]{surfel}{surfel}{surface-element}

\newacronym[shortplural={UAVs}, longplural={unmanned aerial vehicles}]{UAV}{UAV}{unmanned aerial vehicle}

\newacronym{WTA}{WTA}{winner-takes-it-all}

\newacronym{SDF}{SDF}{signed distance function}

\newacronym{TSDF}{TSDF}{truncated signed distance function}
\newacronym{NBV}{NBV}{Next-Best View}
\newacronym{ToF}{ToF}{time-of-flight}

\begin{document}
%
%\title{iVS3D: An Open Source Framework for Intelligent Preprocessing of Image Sequences for 3D Reconstruction }
\title{iVS3D: An Open Source Framework for Intelligent Video Sampling and Preprocessing to Facilitate 3D Reconstruction}

%IF ONLY FOR BLIND SUBMISSION
\iftrue

\author{Max Hermann\inst{1,2} \and
Thomas Pollok\inst{1} \and
Daniel Brommer\inst{1}* \and
Dominic Zahn\inst{1}*
}

\authorrunning{M. Hermann et al.}
\institute{Fraunhofer IOSB, Video Exploitation Systems, Karlsruhe, Germany\\
\email{\{max.hermann,thomas.pollok,daniel.brommer,dominic.zahn\}@iosb.fraunhofer.de} \and
Institute of Photogrammetry and Remote Sensing,
Karlsruhe Institute of Technology, Karlsruhe, Germany\\
\email{max.hermann@kit.edu}\\
}

\fi

\titlerunning{iVS3D: An Open Source Framework for Intelligent Video Sampling and Preprocessing}

\maketitle              % typeset the header of the contribution
%

%IF ONLY FOR BLIND SUBMISSION
\iffalse
\vspace{5em}
\fi

\begin{abstract}
The creation of detailed 3D models is relevant for a wide range of applications such as navigation in three-dimensional space, construction planning or disaster assessment. %
However, the complex processing and long execution time for detailed 3D reconstructions require the original database to be reduced in order to obtain a result in reasonable time. %
In this paper we therefore present our framework iVS3D for intelligent pre-processing of image sequences.  % 
Our software is able to down sample entire videos to a specific frame rate, as well as to resize and crop the individual images. %
Furthermore, thanks to our modular architecture, it is easy to develop and integrate plugins with additional algorithms. %
We provide three plugins as baseline methods that enable an intelligent selection of suitable images and can enrich them with additional information. %
To filter out images affected by motion blur, we developed a plugin that detects these frames and also searches the spatial neighbourhood for suitable images as replacements. %
The second plugin uses optical flow to detect redundant images caused by a temporarily stationary camera.  %
In our experiments, we show how this approach leads to a more balanced image sampling if the camera speed varies, and that excluding such redundant images leads to a time saving of 8.1\percent for our sequences. %
A third plugin makes it possible to exclude challenging image regions from the 3D reconstruction by performing semantic segmentation. %
As we think that the community can greatly benefit from such an approach, we will publish our framework and the developed plugins open source using the MIT licence to allow co-development and easy extension. %

\end{abstract}

\blfootnote{* Contributed equally to this work}
%%%%%%%%%%%%%%%%%%%%%%%%%%%%%%%%%%%%%%%%%%%%%%%
\section{Introduction}
\label{sec:intro}
%%%%%%%%%%%%%%%%%%%%%%%%%%%%%%%%%%%%%%%%%%%%%%%

Photogrammetric approaches like Structure-from-Motion require a set of images in order to reconstruct a 3D scene. In the case of video sequences, not every frame provides additional information that has not been already been available through a previous frame. The manual  preprocessing of image or video sequences can require intensive manual labour, that can be easily automated.
In the case of recorded videos using an drone, there could be situations, where an operator keeps the drone still at a location, before flying to the next location. This means that the input video would contain a large set of redundant frames with very similar image content, while the use of more frames slows down the reconstruction process drastically. Naive approaches, where the original frame rate is down sampled to a lower frame rate, reduce the number of images in the data set drastically, which in turn will have impact on the reconstruction speed as only a subset of images has to be processed in the reconstruction pipeline. However, the naive approach will not always result in a good selection of keyframes. Blurry images impact the image matching and texturing process, where neighbouring frames in the local neighbourhood of the sequence would provide a higher image quality. Also images with a lot of dynamic objects like moving persons can impact the reconstruction quality. Also situations in which the camera is not moving, a selection of a frame with lower individually moving objects or low dynamic static pixel ratio could also contribute to a higher quality. Popular reconstruction approaches like COLMAP \cite{schoenberger2016sfm} allow to provide an additional binary mask per frame in order to prevent mismatches from regions with dynamic objects. Our framework allows to create such masks automatically using a Mask R-CNN \cite{maskrcnn} plugin. In this paper we present our framework iVS3D to tackle the task of intelligent image preprocessing of videos or continuous image sequences for high quality 3D reconstruction purpose. Our framework consists of a plugin based architecture for extensibility. Our framework provides a COLMAP integration and can be extended with further reconstruction tools. The code\setcounter{footnote}{0}\footnote{https://github.com/iVS3D/iVS3D} is released using the MIT open source license. We hope that we can facilitate the task of 3D reconstruction from 2D images with this contribution and invite the community for further contributions and improvements.

The paper is structured as follows: first related work is presented. Afterwards, our approach and architecture is presented in section 3. Experiments using our framework are discussed in section 4 and finally a conclusion is presented.
%%%%%%%%%%%%%%%%%%%%%%%%%%%%%%%%%%%%%%%%%%%%%%%
\section{Related Work}%
\label{sec:related_work}
%%%%%%%%%%%%%%%%%%%%%%%%%%%%%%%%%%%%%%%%%%%%%%%

Pre-processing strategies of video and image sequences for the photogrammetric application of 3D reconstruction can be mainly divided into two categories. The first category tries to reduce the number of frames in favour of a lower computational run-time cost and the second category tries to improve the quality in terms of total 3D point reprojection error and visual quality after texture mapping. The first category is extensively used by and integrated in visual SLAM techniques \cite{9440682} \cite{engel14eccv}, in order to reduce the memory footprint and optimization cost of Bundle Adjustment. New keyframes are sampled from the input sequence only if the relative translation and rotation to the previous keyframe exceeds a manually specified threshold. Additionally, keyframes are removed by these methods during run-time, in case a loop has been detected and the occurrence of redundant keyframes from very similar views. This can result in a better set of keyframes compared to the naive approach, where the complete sequence is simply sampled to a lower frame rate. Bellavia et al. \cite{fastadaptivepreprocessing} proposed an online method to detect and discard bad frames, i.e. blurry frames, resulting in a better reconstruction quality. Ballabeni et al. \cite{advancespreprocessing} propose an image enhancement preprocessing strategy to improve the reconstruction quality. It consists of multiple stages like color balancing, exposure equalization, image denoising, rgb to grayscale conversion and finally an adaptive median filtering. In this paper we do not propose a new preprocessing strategy. The main contribution of our paper is the contribution of an extensible open source framework with a number of baseline plugins to the community, for preprocessing of image and video sequences in the context of photogrammetric applications.

\clearpage

%%%%%%%%%%%%%%%%%%%%%%%%%%%%%%%%%%%%%%%%%%%%%%%
\section{Approach}%
\label{sec:methodology}
%%%%%%%%%%%%%%%%%%%%%%%%%%%%%%%%%%%%%%%%%%%%%%%

Our framework has two main goals: One is to speed up the downstream 3D reconstruction by filtering images without enough novel image content while maintaining the reconstruction quality. %
Secondly, the enhancement of the source material by removing challenging areas. %
In this context, we mainly focus on masking regions with a high degree of movement, such as pedestrians, cars or vegetation. %
For easy extensibility and quick addition of new algorithms, we have decided to use a plugin-based architecture. %
All of our baseline methods presented below are encapsulated as  plugins and can thus be individually added to the processing chain. %
In the following, we will first discuss the basic architecture of our framework and then describe the individual plugins with regard to frame filtering and image region masking. %
Finally, we outline our ability to define, save and execute specific workflows for batch processing. %

\subsection{Architecture overview}
We have designed the application as a model-view-controller architecture that can integrate algorithms via two distinct interfaces. %
iVS3D is implemented with the Qt framework using the programming language C++ and is therefore cross platform capable. %
To allow everyone to add algorithms to iVS3D and to easily share them we are using a plugin-based approach for all the algorithms. %
Plugins can either extract specific images from the provided image sequences or generate additional image information. %
These explicitly selected images are called keyframes in the following. %
For this architecture we rely on the plug-in functionality in Qt. %
As baseline methods we implemented four plugins, which are the Nth frame plugin for fixed frame subsampling, the camera movement detection plugin, the blur detection plugin and the plugin for semantic segmentation. %
%Namely we start with a import (\textbf{1}) then we use the plugins to sample images (\textbf{2}) and lastly the sampled images are exported (\textbf{3}).
%\begin{figure}[h]
%    \centering
%    \includegraphics[width=\textwidth]{figures/architecture/Architecture_Overview.png}
%    \caption{Architecture is based on a MVC-concept with the option to add two types of plugins. Firstly there are plugins to select keyframes, while the second type of plugins offers enriched data.}
%    \label{fig:architecture_overview}
%\end{figure}

\begin{figure}[h!]
    \centering
    \includegraphics[width=\textwidth]{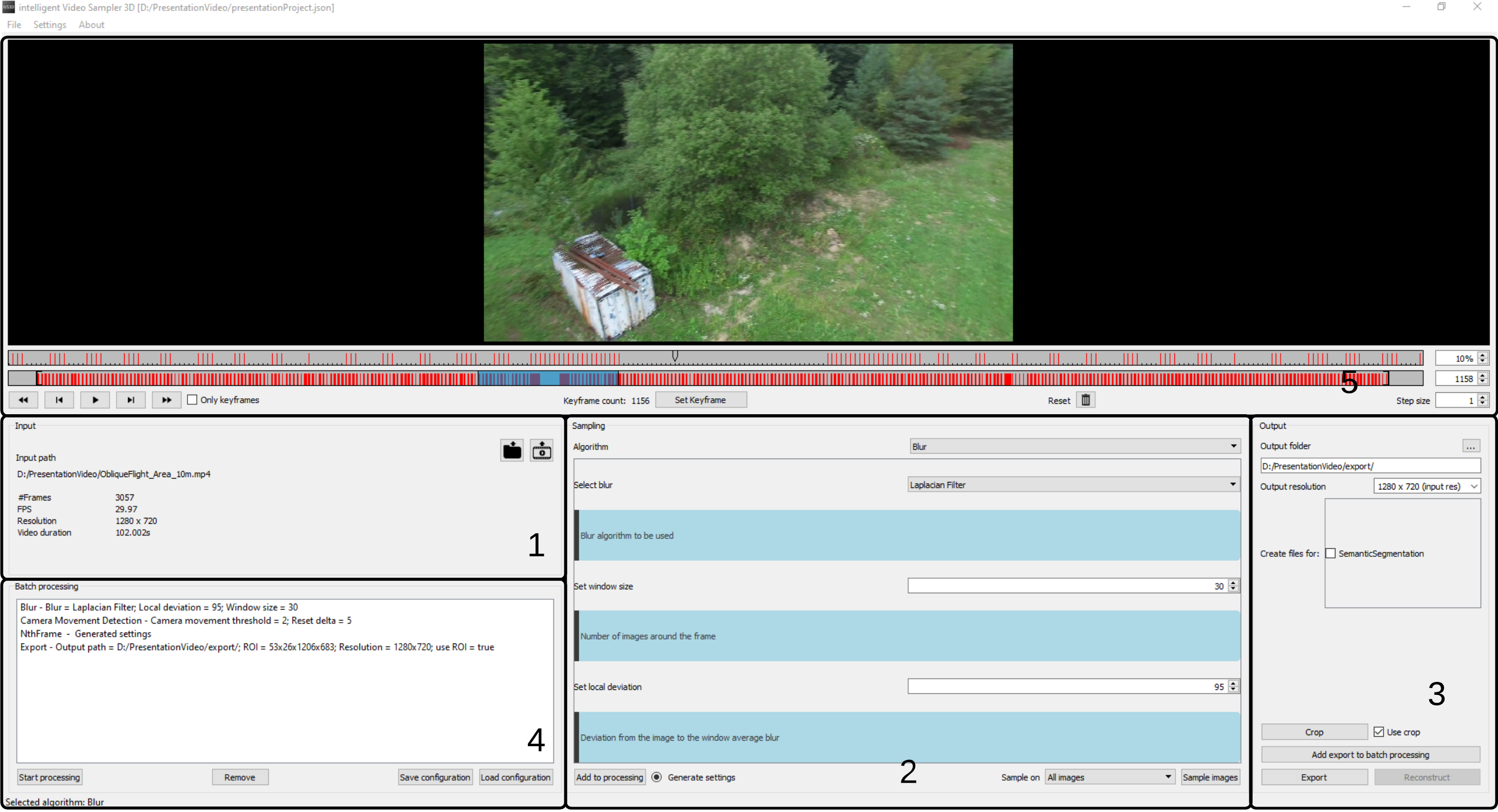}
    \caption{Graphical user interface which is split in five different sections. 1. Input, 2. Plugins, 3. Export, 4. Batch processing and 5. Video player with the timeline for keyframes}
    \label{fig:gui}
\end{figure}

The graphical user interface shown in \Cref{fig:gui} gives an impression of the individual components of our framework, which we will discuss in more detail in the following. %
Our processing pipeline consists of a defined input that receives either a video or a collection of individual images. %
These are transformed by one or more plugins and then exported to a new location. %
In addition, the software COLMAP can be started directly from our application with automatically generated project files. %
%Each step in the processing pipeline is logged in detail in a JSON file so that the executed steps can be traced in retrospect. %
In the following, the individual steps of the pipeline are explained in more detail, whereby the sections are structured according to the components marked in \Cref{fig:gui}. %

\subsubsection{1 Input} 
To start the process, you can either open a project that is already in progress or begin importing new data. %
For this we accept a folder with single images as well as videos in the common formats. %
The project file contains the source of the imported data as well as the settings and sampled key frames. %
If a video file was imported the graphical user interface will display additional information in the input section like the frames per second, resolution and duration. %

\subsubsection{2 Plugins}
To process the input data we offer two types of plugins. %
The first one represents the possibility to select keyframes from the currently loaded image sequence, while the second one generates additional information for each image. %
In order to enable the user to set parameters for the plugin, every plugin has its own user interface where all necessary settings can be applied, which gets embedded in the main application. %
For more complex plugins with several parameters, we offer an optional interface that allows the plugin to use heuristics to determine suitable settings in advance based on a subset of the input data. %
This allows, for example, algorithms whose parameters depend indirectly on the image resolution to set more suitable default parameters. %
In contrast to this there are some functions that need to be implemented to ensure that the plugin will work correctly. %
The \textit{sampleImages} function is the core of each plugin. %
It receives access to the input data and returns a vector of indices of keyframes based on the given image sequence. %
To start the sampling process a plugin has to be selected and configured by the user. %
With the brackets in the timeline \textbf{5} the amount of images which are sampled can be reduced by excluding images on the respective left or right side. %
In this way, parts of videos can be removed that show, for example, a starting or landing \gls*{UAV}. %
After the sampling is done the timeline will show the sampled key frames as red bars. %
Using multiple sampling processes in sequence is supported as well as manual selecting and removing keyframes. %
%Plugins have the possibility to buffer values after they finished the sampling process. %
%When available this buffer will be used in the plugin which can significantly lower the time it takes the sampling to finish. %
%These buffers are saved in the project file and can be used any time later on the same input. %
The second type of plugin focuses on creating additional information that belong to a specific input image. %
Like the plugins of the first type, these plugins have their own user interface, which is embedded in the main window when the plugin is selected. %
The core of these plugins is the \textit{transform} function where they are given a image and return a vector of images. %
It has the ability to show the newly generated images on the graphical user interface and to export the newly generated images among the sampled images. %
A deeper explanation of the already implemented plugins is given in \Cref{subsec:filtering_images}. %

\subsubsection{3 Export}
The last step of the processing pipeline is to export the sampled images to the desired location. %
In addition to the export of the sampled images, the export step offers additional functionality. %
Images can be cropped to e.g. remove unwanted timestamps visible in the images and the resolution can be changed. %
Every export creates a project file containing the import source, the indexes of the sampled keyframes. %
After an export is finished a 3D reconstruction tool can be started from iVS3D. %
So far, we have integrated the software COLMAP and generate project file and database. %
If binary masks for the images were created these will be included to the project file as well. %
COLMAP then starts with these files already imported which allows the user to start the reconstruction process without having to manually set the input data again. %
Furthermore, the direct start of a headless reconstruction with COLMAP is possible with our application. %

\subsection{Baseline plugins already developed} \label{subsec:filtering_images}
The time complexity of methods such as feature matching or structure from motion often scale quadratic, which makes processing large image data sets very difficult. %
Choosing the right images is therefore crucial to avoid unnecessary processing time. % 
We have implemented three filtering plugins as baseline methods. %
Firstly, a simple down-sampling to a fixed fps value, then the detection of redundant frames by determining the optical flow and thirdly the filtering of blurred images. %
We have also developed a plugin that uses semantic segmentation to exclude regions from the image that are unusable for 3D reconstruction. %

\subsubsection{Plugin to exclude redundant images}
If the images for the 3D reconstruction are extracted from a video, it can happen that the camera is stationary for some time and therefore the following images are nearly identical. %
To detect and filter these redundant images, we identify missing camera movement by calculating the optical flow following the method of Farneb{\"a}ck\cite{farneback2003two} and specify a threshold for minimum camera movement. %
%\vspace{10pt}
%\begin{algorithm}
%\caption{Filter based on camera movement}
%\begin{algorithmic}
%\Require $inputFrames.isOrdered$ \& $resetDelta > 0$ \& $movementThreshold > 0$
%\State $referenceFrame := 0$
%\State $resetMovement := 0$
%\For{$currentFrame : inputFrames$}
%\State $optFlowMat := calculateFlow()$
%\State movement \leftarrow$ movementFromMatrix($optFlowMat$)$ + resetMovement$
%\If{movement > movementThreshold$}
%\State $keyframes$.add($currentFrame$)
%\State $referenceFrame \leftarrow currentFrame$
%\ElsIf{$currentFrame > currentFrame + resetDelta$}
%\State $referenceFrame \leftarrow currentFrame$
%\State $resetMovement \leftarrow movement$
%\EndIf
%\EndFor
%\end{algorithmic}
%\end{algorithm}
The total algorithm can be split in three major steps: 1.) Calculate optical flow with the method of Farneb{\"a}ck 2.) Estimate movement of the camera 3.) Select keyframes. %
As mentioned before the plugin calculates the optical flow between two frames using the OpenCV\cite{opencv_library} implementation of the Farneb{\"a}ck algorithm. %
Because this step can take some time we added the possibility to use a CUDA optimised variant. %
The first frame is called the reference image because it will be the starting point to which every following image will be compared to until a new keyframe is selected. %
%For this we use mostly standard parameter with some little changes to improve the speed of the algorithm while still generating a good enough approximation. %
%Because of the approximate nature and the already underlying noise through moving persons, cars or other objects we can sacrifice precision to gain computation speed without losing important information. \\
From the estimated values of the optical flow we calculate the trimmed mean to get a score for the amount of movement between two images, which is compared to the parameter \textit{movement threshold}. %
This approach is methodologically difficult if two images are compared that do not show sufficient similarity. %
To prevent this from happening the plugin uses a second parameter named \textit{reset delta}. %
It determines how many frames are allowed to be between two compared images until in each case a new reference frame is set and the process starts again from the beginning. %
In contrast to approaches such as \gls*{SLAM}, we do not need any further information in addition to the images. % 
This includes the fact that a calibrated camera is not required to use this plugin. %

\subsubsection{Plugin to exclude images affected by motion blur}
Sudden movements can lead to motion blur, which significantly reduces the quality of such images. %
Using a sliding window approach, we search for images without motion blur and, if motion blur is detected, for spatially close replacement images. %
We calculate a sharpness score for each image and compare it with the values of surrounding images to get a relative score for the sharpness of the image. %
To calculate this score, we rely on established methods. %
Pertuz et al. state in their work that depending on the image content, type of camera and level of noise, different algorithms produce competitive results \cite{focus_measure_operators}. %
We therefore offer two options as baseline methods. %
First, the Tenegrad algorithm using Sobel operators according to the implementation described in \cite{focus_measure_operators} and second a technique utilising the variance of the image Laplacian \cite{pech2000diatom}. %
If the sharpness score of an image divided by the average value of its window is bigger than a defined threshold, this image is considered sharp and will be selected as keyframe. %
%\begin{algorithm}
%    \caption{Filter blurry images}\label{algo:blurry_images}
%    \begin{algorithmic}
%    \Require $int$ $windowSize, double$ $localDeviation, vector<double>$ $blur$
%    \State $result \longleftarrow new$ $Vector<int>$
%    \For{$i = 0$ $to$ $blur.size$}
%    \State $neighbourValues \longleftarrow  [blur.get(i - windowSize),.., blur.get(i + windowSize)]$
%    \State $average \longleftarrow neighbourValues.avg$
%    \If{$localDeviation > (blur.get(i)/average) * 100$}
%    \State $result.add(i)$
%    \EndIf
%    \EndFor
%    \State \textbf{return} $result$
%    \end{algorithmic}
%\end{algorithm}
In case keyframes are already sampled, we can try to obtain keyframes with a lower motion blur by taking the neighbouring images into account. %
A window of a certain size is created around each keyframe, in which the image with the highest sharpness value is selected as the substitute keyframe. % 
%\begin{algorithm}
%    \caption{Move keyframes}\label{algo:move_frames}
%    \begin{algorithmic}
%    \Require $int$ $windowSize, vector<double>$ $blur, vector<int>$ $keyframes$
%    \State $result \longleftarrow new$ $Vector<int>$
%    \For{$i = 0$ $to$ $keyframes.size$}
%    \State $currentImage = keyframes.get(i)$
%    \State $neighbourValues \longleftarrow  [blur.get(currentImage - windowSize),.., blur.get(currentImage + windowSize)]$
%    \State $maxImage \longleftarrow neighbourValues.argmax$
%    \State $result.add(maxImage)$
%    \EndFor
%    \State \textbf{return} $result$
%    \end{algorithmic}
%\end{algorithm}
%\begin{itemize}
%    \item sliding window approach
%    \item selects neighbouring frames instead
%    \item technique?
%    \item pseudocode?
%\end{itemize}

\subsubsection{Enrichment with additional information by semantic segmentation}\label{sec:masking}
Objects such as pedestrians or cars that move dynamically in the scene hinder 3D reconstruction because it is more difficult to find correspondences between images. %
Instead of selecting keyframes as previously explained, this plugin generates additional information by masking the challenging areas in the input images by incorporating semantic information. %
In this way we prevent not only unstable matches of feature points between images but also that these regions have to be processed at all. %
For this we use the ability of COLMAP to use binary masks as input in addition to the images. %
We generate these masks through semantic segmentation by transferring the included classes into the binary mask. %
For our baseline plugin we use neural networks consisting of a DeepLabv3+ \cite{deeplLab} model with a ResNet101 \cite{he2016deep} backbone. %
For a faster execution speed on weaker hardware, we provide networks trained on different resolutions. %
All networks are trained on the Cityscapes data set \cite{cordts2016cityscapes} and provide segmentations for 19 classes. %
However, it is very easy to add additional neural networks with different architectures or which have been trained on other data sets. %
Our developed plugin is able to display a live preview of the segmentations as well as the binary mask. %
In this way, combinations of classes or different deep learning models can be tested quickly. %
For broad compatibility we use OpenCV as inference framework which allows a fallback to the CPU if no CUDA compatible graphics card is available. %

\subsection{Batch Processing}
To prevent the user from having to perform all the steps described above for every image sequence batch processing can be used. %
Every sampling plugin can be added multiples times with any configuration wanted. %
Likewise, multiple export steps with different settings can be added to the workflow. %
Multiple exports can be useful to e.g. export images with different resolutions or with varying frame rate to different locations. %
The currently selected workflow can be seen and edited in \textbf{4} on the graphical user interface shown in \Cref{fig:gui}. %
Starting the batch processing will run every step in the workflow in the defined order. %
The specified workflow can be saved and loaded into other projects, and also used to start our software in headless mode. % 
In this case, an input sequence must be specified for which batch processing is then performed. %
%%%%%%%%%%%%%%%%%%%%%%%%%%%%%%%%%%%%%%%%%%%%%%%
\section{Experiments}
\label{sec:eval}
%%%%%%%%%%%%%%%%%%%%%%%%%%%%%%%%%%%%%%%%%%%%%%%

\begin{figure}[h!]
    \centering
    \includegraphics[width=0.2\textwidth]{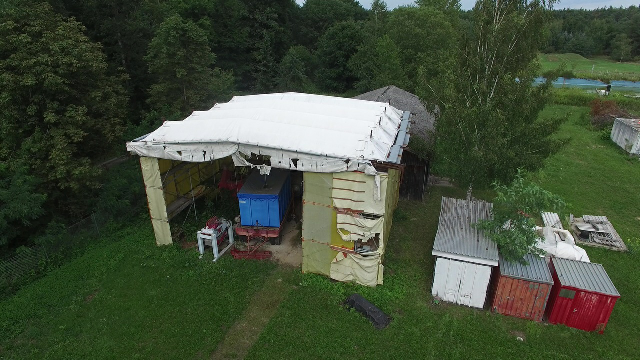}%
    \includegraphics[width=0.2\textwidth]{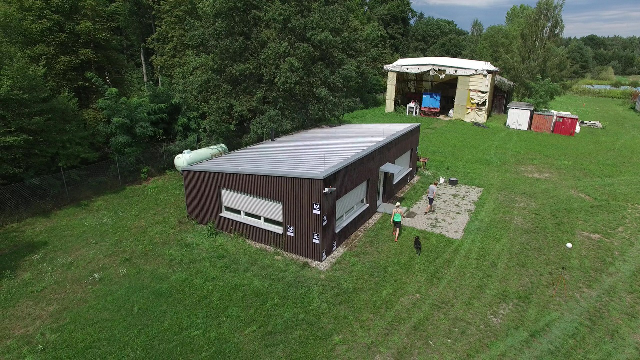}%
    \includegraphics[width=0.2\textwidth]{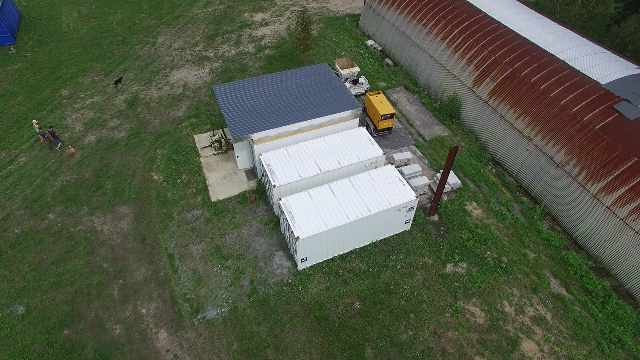}%
    \includegraphics[width=0.2\textwidth]{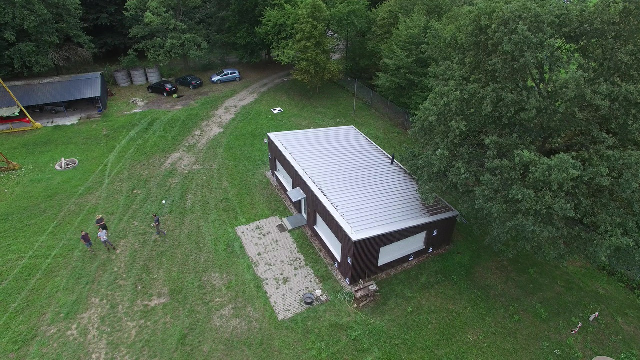}%
    \includegraphics[width=0.2\textwidth]{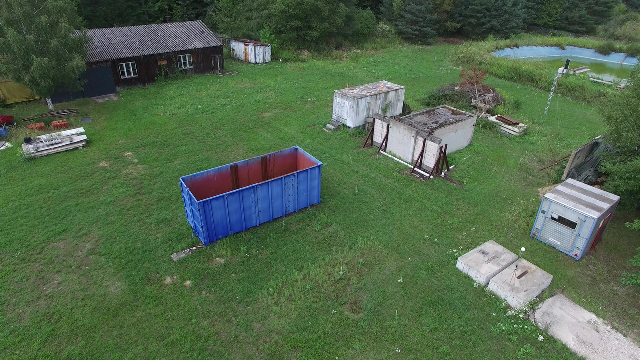}\\%
    \vspace{-1mm}
    \includegraphics[width=0.2\textwidth]{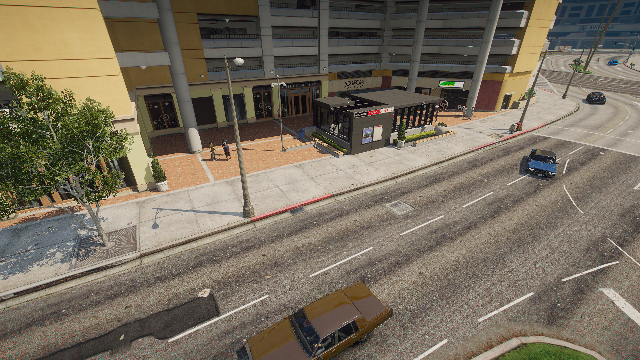}%
    \includegraphics[width=0.2\textwidth]{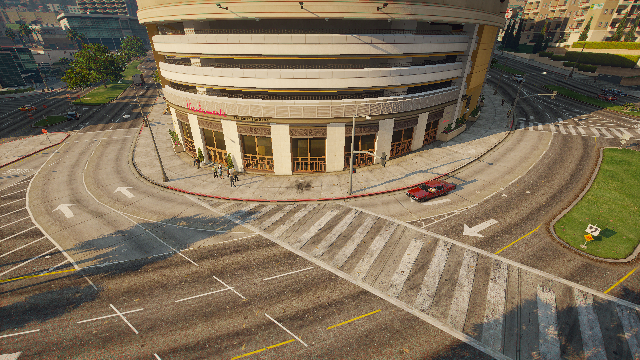}%
    \includegraphics[width=0.2\textwidth]{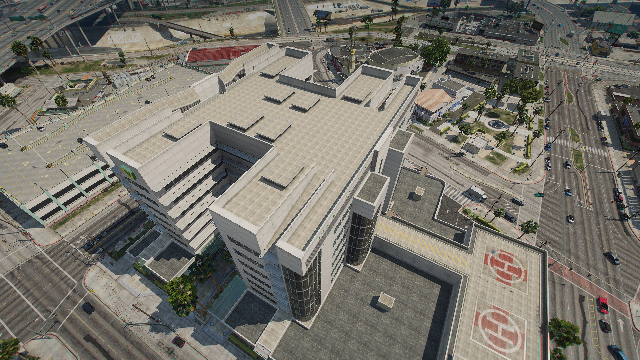}%
    \includegraphics[width=0.2\textwidth]{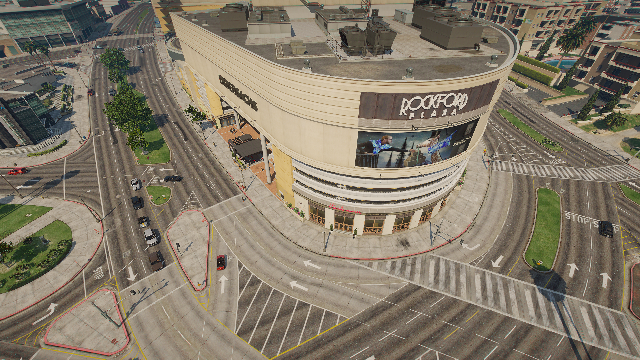}%
    \includegraphics[width=0.2\textwidth]{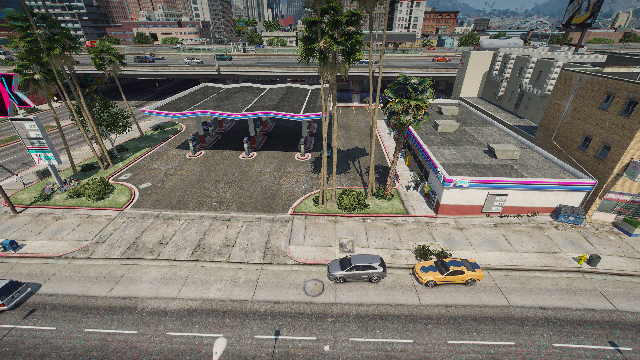}\\
    \caption{The first row shows examples of the rural real-world data set and the second row from the urban synthetic data set. %
    The sequences containing the last three images of the second row are referred to as A, B and C in the following evaluation. %
    }
    \label{fig:examples_dataset}
\end{figure}

To evaluate the advantage of our framework, we compare the output with the widely used approach of sub-sampling the input image sequences to a fixed fps rate. %
We will first focus on the evaluation of our plugins for frame selection and to what extent they have an influence on the quality and duration of the 3D reconstruction. %
In addition, the following section examines if the quality of the reconstructions can be further improved by removing challenging image regions. %
For this, complete 3D reconstructions are created with COLMAP for the sequences preprocessed by our framework and a corresponding 1\fps baseline. %
Our video data sets for this come from two different sources. %
The first consists of aerial imagery taken from oblique view by an \gls*{UAV} flying around a point of interest. %
This scenario features a rural setting where buildings and objects were captured at different altitudes and camera angles by a DJI Phantom 3 Pro. %
The flight altitude varies across the sequences from 2\m to 18\m and cars as well as moving people are visible in some scenes. %
They also include motion blur and scenes where the \gls*{UAV} is stationary. %
Due to the lack of ground truth data, we can evaluate the achieved results here mainly qualitatively and measure the speed increase of the reconstruction by excluding redundant images. % 
In order to quantitatively evaluate the quality of the estimated camera trajectories and depth maps from COLMAP, we therefore additionally recorded synthetic sequences.  %
We use a modification of the code from Johnson-Roberson et al. \cite{johnson2016driving} to gather data from the video game GTAV. %
This allows us to extract ground truth depth maps and camera trajectories from the GPU. %
Unfortunately, there is no way to access the underlying 3D models and simple back projection of the depth maps only works if the scene does not contain moving objects, otherwise artefacts will occur. %
For the evaluation, we recorded eight synthetic video sequences and their ground truth, which contain areas with a lot of movement and sections where the flight speed of the \gls*{UAV} varies. %
The synthetic sequences differ from the real-world data set mainly by the urban setting and a partly higher flight altitude. %
Examples of the two data sets are visible in \Cref{fig:examples_dataset}. %

\subsection{Experimental setup for the frame selection}
For our frame selection we rely on the following three plugins: 1) First we filter out redundant frames 2) Then we down sample the video to 1\fps 3) In the last step we look for blurred images and try to replace them with better neighbouring frames. %
To evaluate this, we perform dense COLMAP reconstructions at both full frame rate as well as at 1\fps and using our approach described above. %
All reconstructions were carried out on a Nvidia Tesla P40 with a Xeon E5-2650 CPU. %
To assess the performance of our framework, we compare the results achieved with COLMAP to the ground truth of our synthetic data set. %
As a metric for evaluating the quality of our depth maps in this context, we use the accuracy $\delta_{\theta}$, which is defined as follows:
\begin{equation}
    \delta_{\theta}\left( d, \hat{d} \right) =\frac{ 1 }{ m }\sum^{m}_{i=1}{ \max\left(\frac{ d_{i} }{\hat{d}_{i}  }, \frac{ \hat{d}_{i} }{d_{i}}\right) < \theta .}
\end{equation}
It classifies a pixel in the estimated depth map as correct if the estimate is within a certain threshold $\theta$ to the corresponding measurement. %
$\delta_{1.25}$ for example describes the proportion of pixels, relative to the number of pixels $m$, for which an estimate exists and for which the difference between the estimate $d$ and the ground truth $\hat{d}$ is not greater than $25$\% of $\hat{d}$. %
For the evaluation of the camera trajectory, we use the \gls*{RMSE} to measure the deviation of the estimated camera position from the ground truth pose. %
We use the library evo \cite{grupp2017evo} for the quantitative evaluation and visualisation of the camera trajectories. %

\begin{figure}[h!]
    \centering
    \small Sequence A~~~~~~~~~~~~~~~~~~~~~~~~~~~~~Sequence B~~~~~~~~~~~~~~~~~~~~~~~~~~~~~Sequence C\\
    \vspace{-0.1em}%
     \rotatebox{90}{\small~~~~~~~~~~~~~~1\fps sampling }%
     \hspace{0.1em}%
        \includegraphics[trim={0.6cm 0.7cm 0.7cm 0},clip,width=0.32\textwidth]{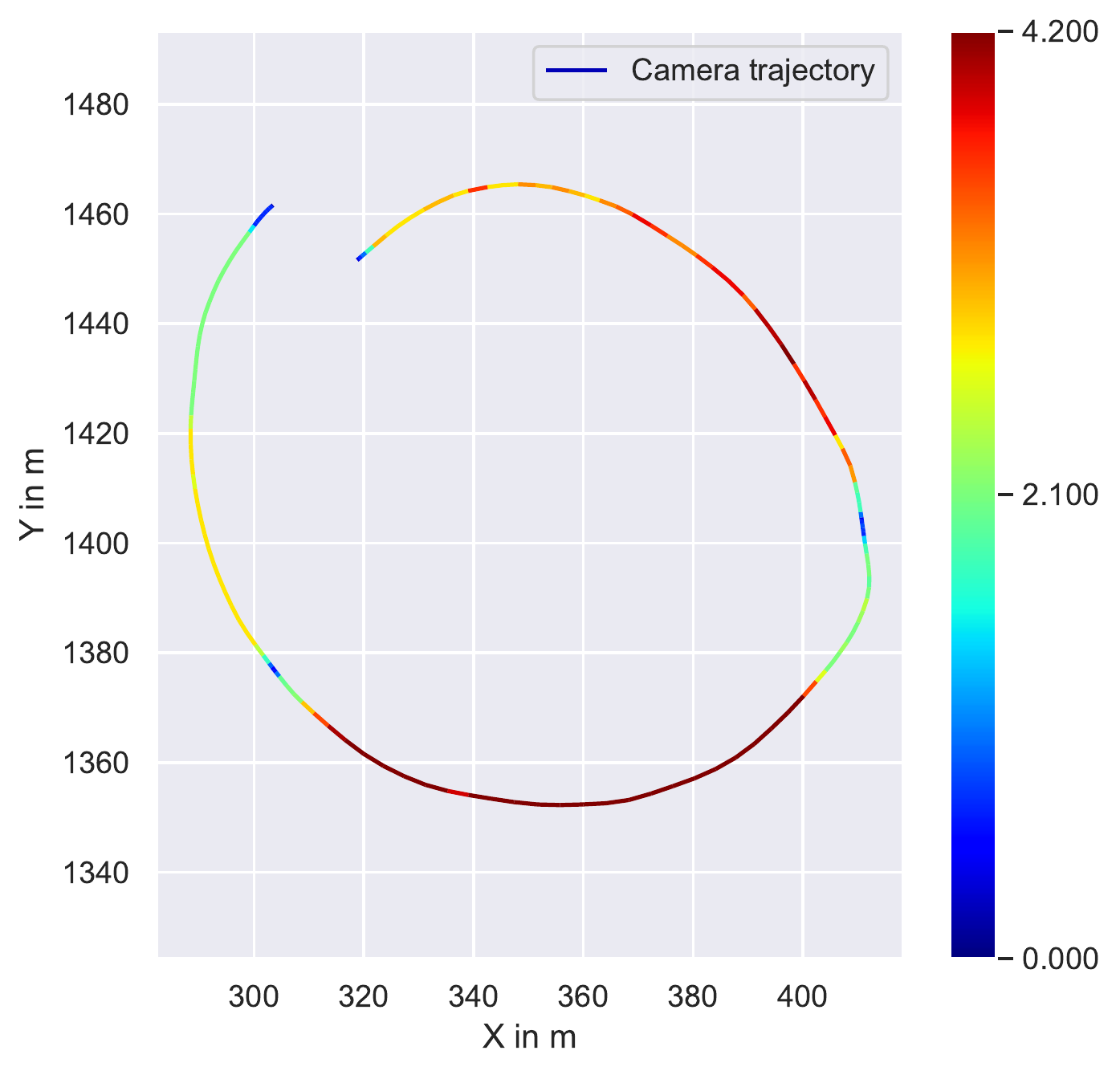}%
        \includegraphics[trim={0.6cm 0.7cm 0.7cm 0},clip,width=0.32\textwidth]{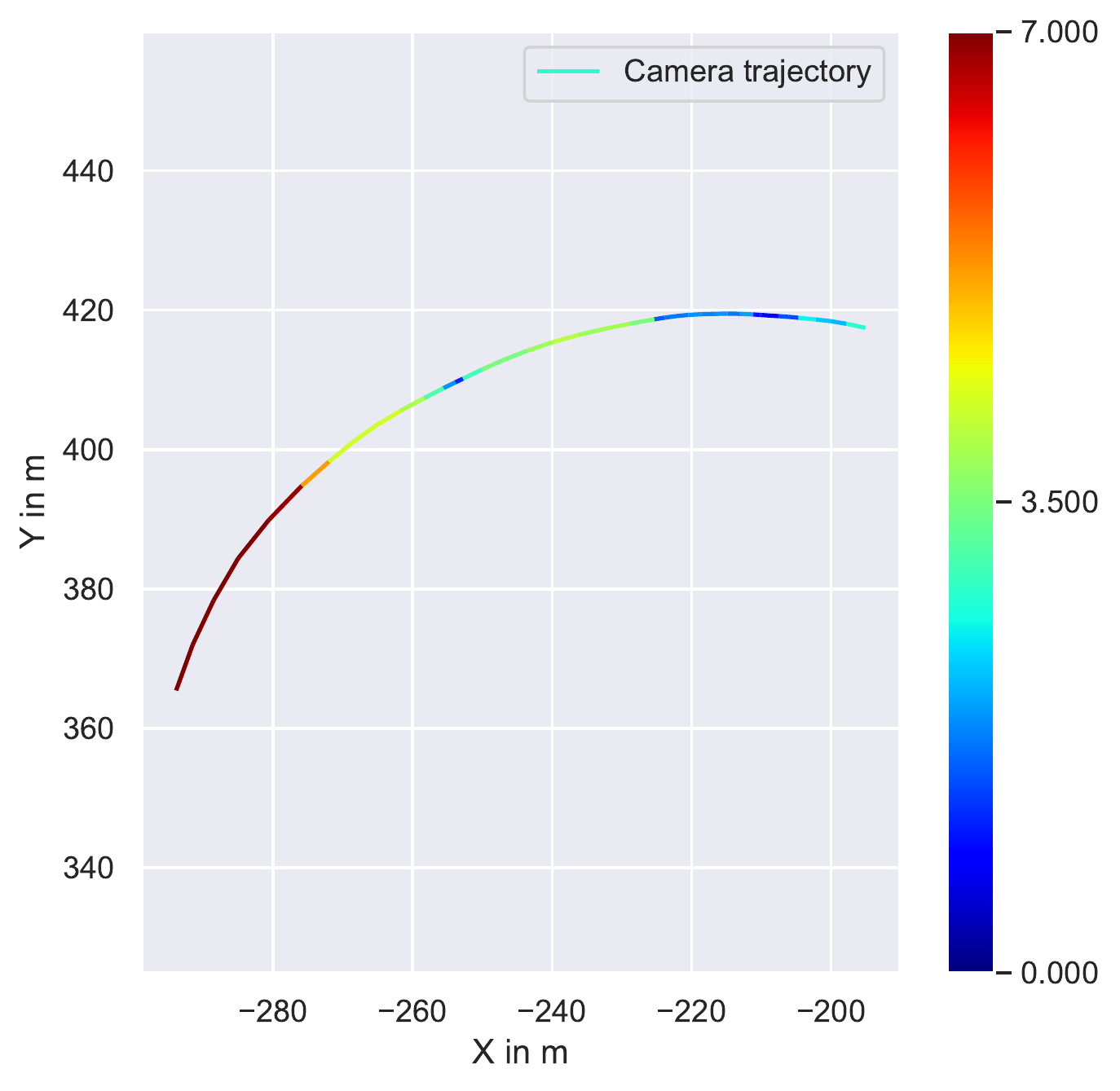}%
        \includegraphics[trim={0.6cm 0.7cm 0.7cm 0},clip,width=0.32\textwidth]{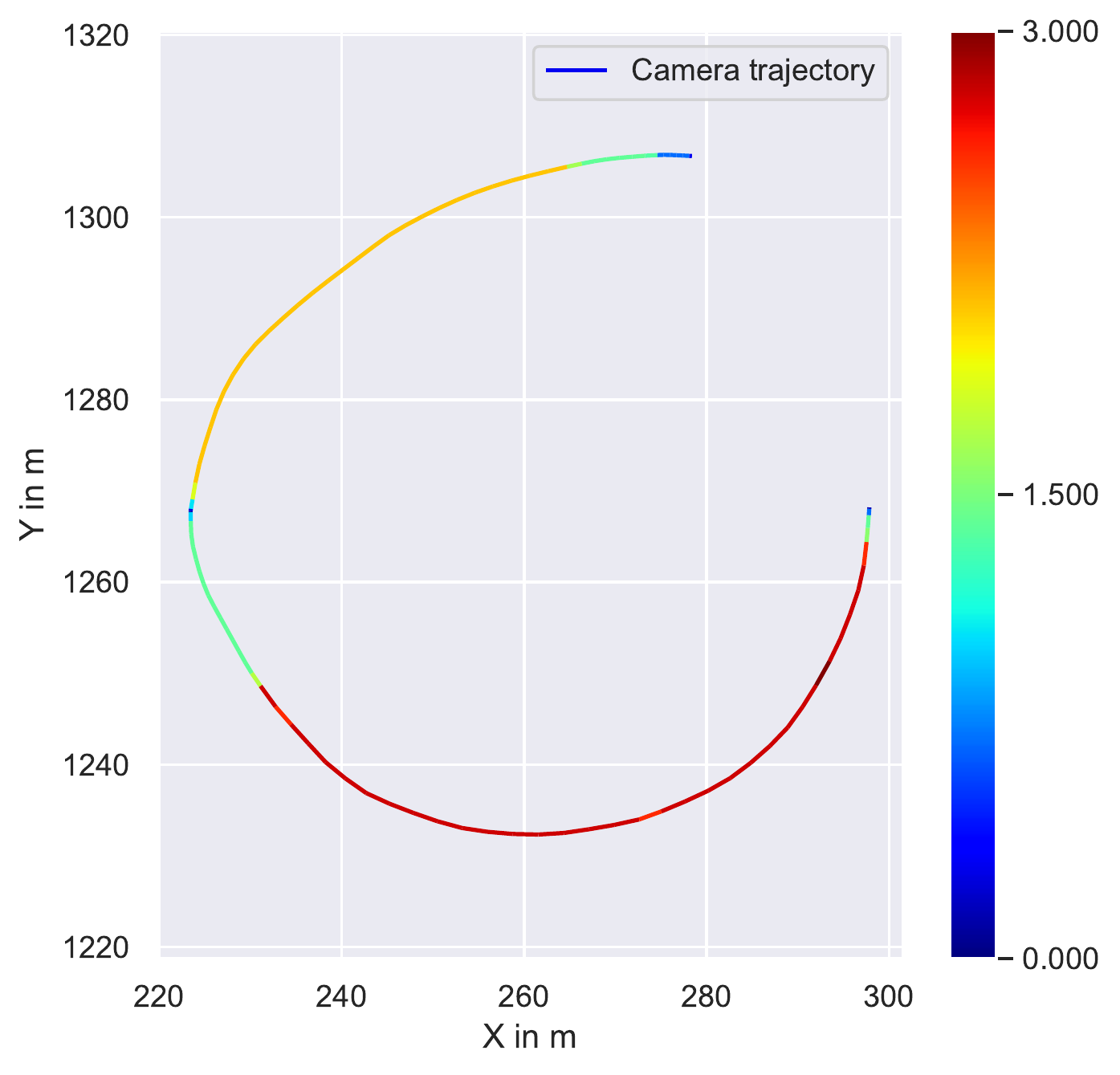}\\
        \vspace{-0.2em}%
        \rotatebox{90}{\small~~~~~~~Sampling with optical flow }%
         \hspace{0.1em}%
        \includegraphics[trim={0.6cm 0.7cm 0.7cm 0},clip,width=0.32\textwidth]{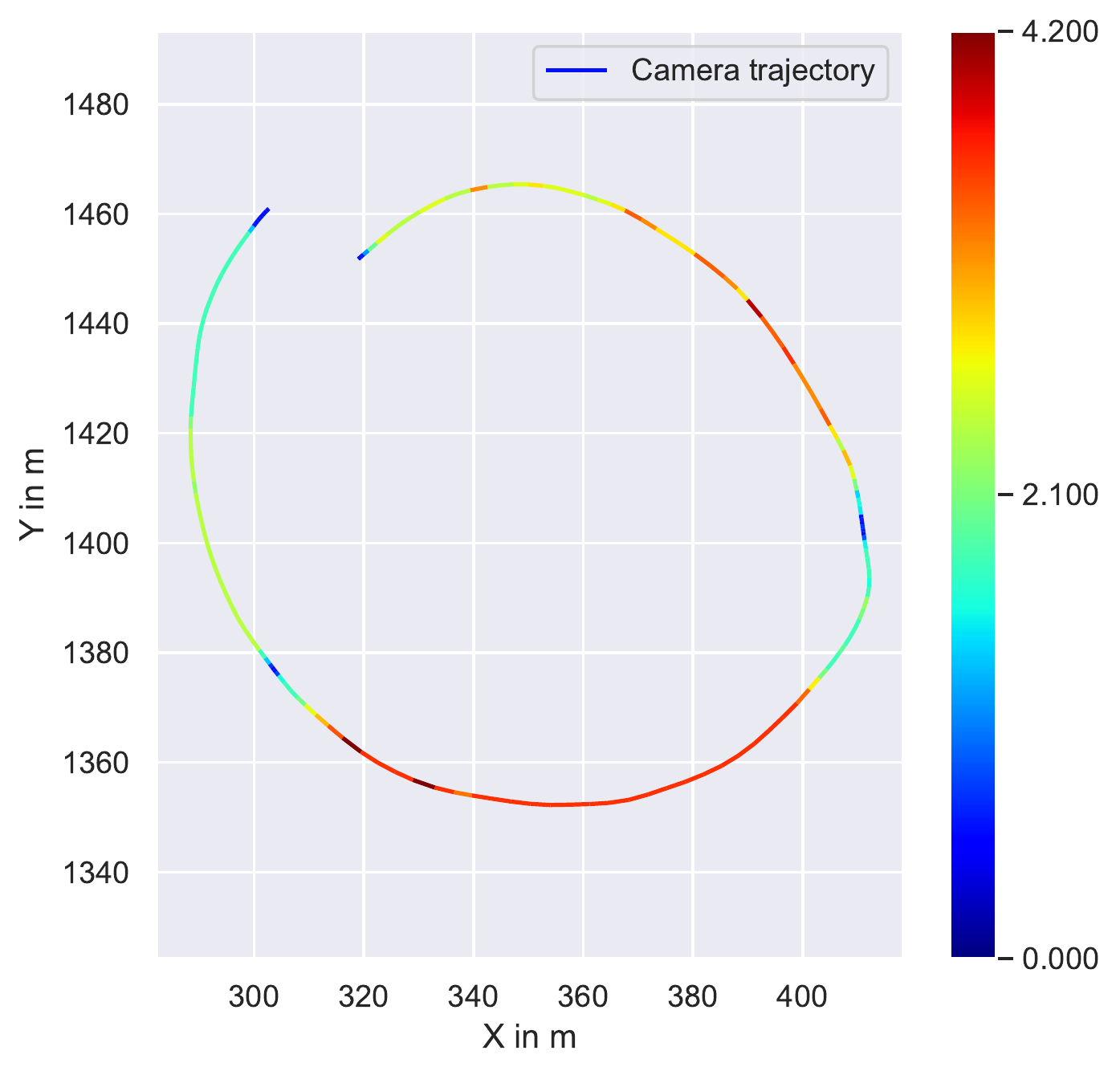}%
        \includegraphics[trim={0.6cm 0.7cm 0.7cm 0},clip,width=0.32\textwidth]{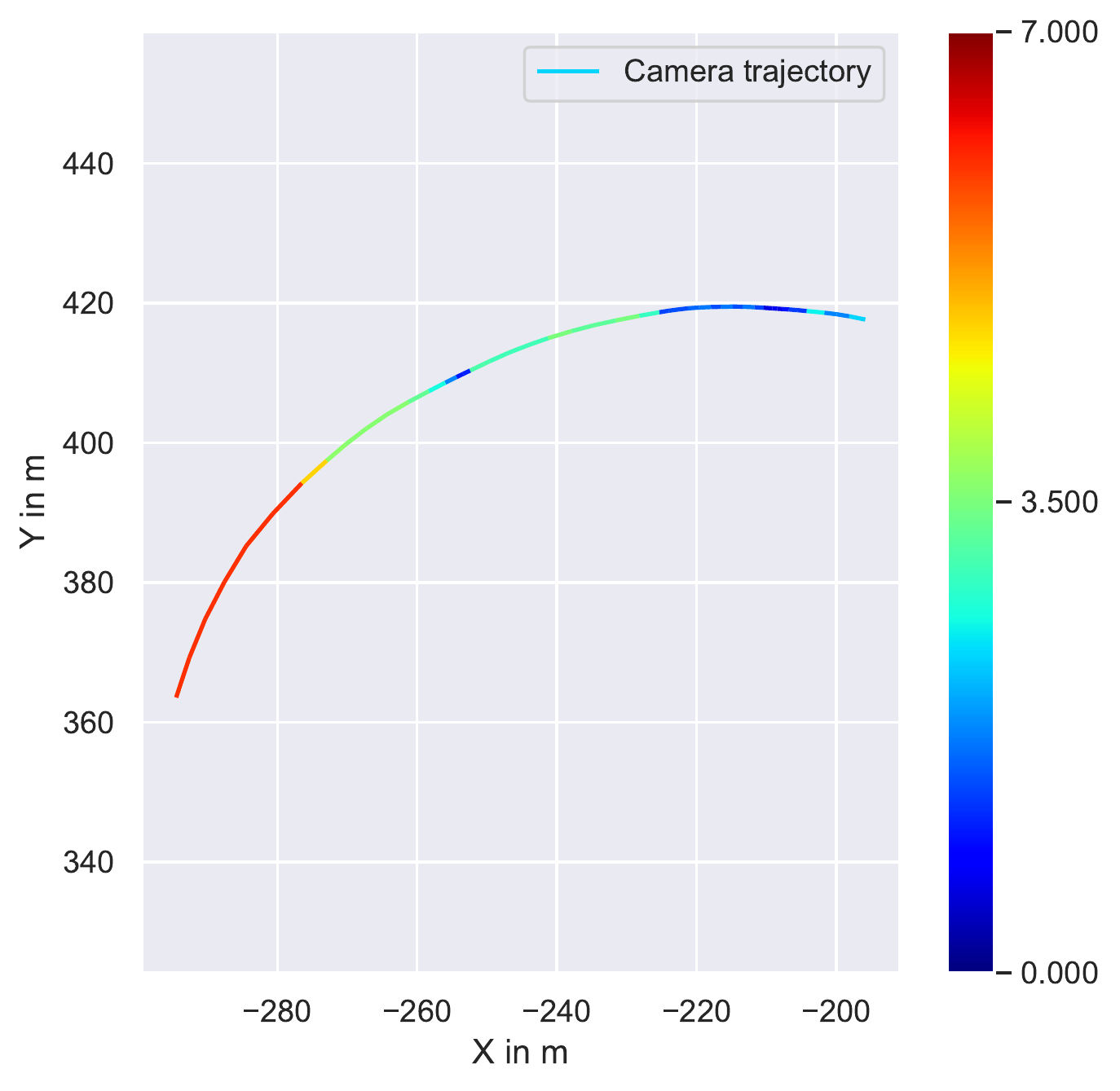}%
        \includegraphics[trim={0.6cm 0.7cm 0.7cm 0},clip,width=0.32\textwidth]{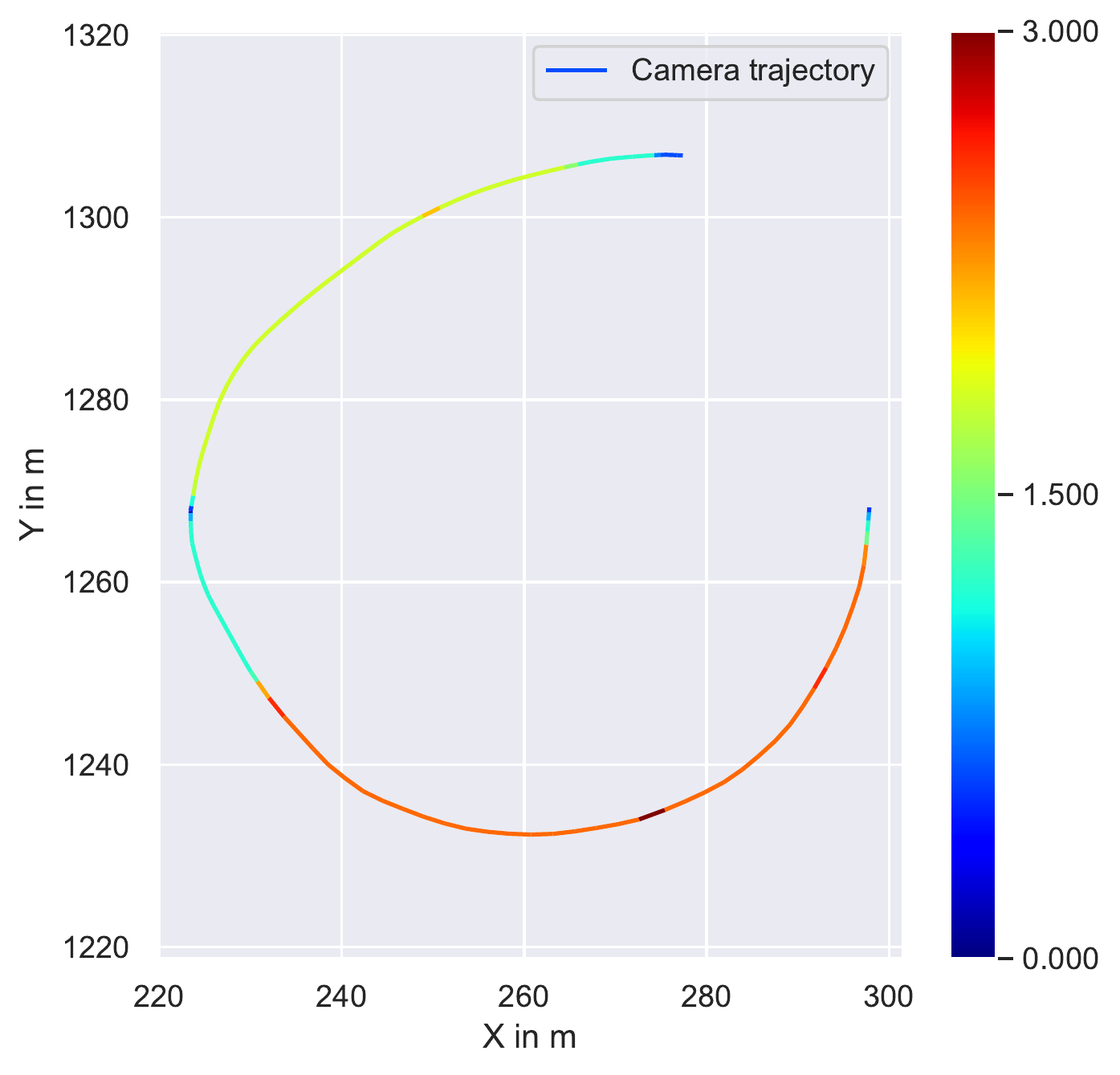}\\
    \caption{Exemplary comparison of the three trajectories sampled in the first row using 1\fps and in the second row using optical flow. %
    The trajectories show the flight path of a \gls*{UAV} in metres, projected onto the $XY$-plane, with the distance to the previous image position colour-coded.
    %Regions in which the distances between the images are large due to high flight speeds are marked in dark red and sections in which the camera is stationary are marked in dark blue. %
    Especially in regions with high flight speed it is visible that our sampling using the optical flow is more balanced. %
    }
    \label{fig:trajectories}
\end{figure}

\begin{table}[h!]
    \centering
    \setlength\tabcolsep{6pt} % default value: 6pt
    \setlength{\extrarowheight}{6pt}

    \begin{tabularx}{0.935\textwidth} {|c|c|c|c|c|c|c|}
        \hline
        \rule{0pt}{\normalbaselineskip}%

        Sequence&Sampling method&$\sigma$&$Q_{0.1}$&$Q_{0.9}$&\gls*{RMSE} Pose&$\delta_{0.05}$ Accuracy\\
        \hline
        \multirow{2}{*}{A}&1\fps&$1.26$\m&$0.70$\m&$4.20$\m&0.0329\m&0.9417\percent\\
         \cline{2-7} 
         &optical flow&$0.95$\m&$0.65$\m&$3.60$\m&0.0296\m&0.9420\percent\\
        \hline
          \multirow{2}{*}{B}&1\fps&$0.90$\m&$0.70$\m&$2.80$\m&0.0066\m&0.9596\percent\\
         \cline{2-7} 
         &optical flow&$0.64$\m&$0.90$\m&$2.40$\m&0.0052\m&0.9594\percent\\
        \hline
          \multirow{2}{*}{C}&1\fps&$1.95$\m&$0.80$\m&$6.68$\m&0.0245\m&0.8886\percent\\
         \cline{2-7} 
         &optical flow&$1.62$\m&$1.03$\m&$6.00$\m&0.0311\m&0.8894\percent\\
        \hline

    \end{tabularx}%
    \vspace{6pt}
    \caption{Quantitative results for three sequences reconstructed with COLMAP, once using 1\fps as well as using optical flow based sampling. %
     In almost all cases, the 10\percent percentile is higher and the 90\percent percentile is lower than the baseline of the 1\fps sampling. %
    Furthermore, all three sequences have a lower standard deviation with our approach. %
    The two columns on the right show that our approach does not lead to inferior quality, as the errors for pose and depth estimation indicate. %
}
    \label{tab:quantitative_results}
\end{table}

Both quantitatively and qualitatively from the colouring of the trajectories in \Cref{fig:trajectories}, it can be seen that for most cases the sampling through our framework leads to a more balanced camera trajectory than sampling based on frame rate alone. %
As shown in \Cref{tab:quantitative_results}, the 10\percent percentile of the distance to the previous camera position is almost everywhere above and the 90\percent percentile below the baseline method of 1\fps sampling. %
In addition, the standard deviation of our sampling approach through the optical flow is noticeably lower. %
This shows that not only are fewer images sampled in regions with little camera movement, but also that more frames are selected in sections with a lot of camera movement. %
As visible in \Cref{fig:trajectories}, not all images where the camera is stationary are excluded. We suspect that this is due to some moving objects within the scene which may prevent these images from being marked as redundant. %
In our experiments, our approach has no negative impact on the quality of the camera trajectories and depth maps. %
As shown in \cref{tab:quantitative_results}, the differences are small and can be explained by run time variability. %
However, since the resulting point clouds contain on average 5.76\percent more points, we assume that the more spatially homogeneous sampling provides better coverage of the 3D scene than selecting images with a fixed frame rate. %
In order to verify this assumption, a comparison of the resulting 3D reconstructions with ground truth 3D models would have to be carried out. %
Unfortunately, to our knowledge, there are no benchmark data sets with 3D models for aerial imagery from an oblique view, which is probably due to the difficult data acquisition. %

In contrast to this approach, however, sampling through optical flow can also be used to reduce the amount of data to be processed by excluding redundant images. %
As before, sections in which the \gls*{UAV} is stationary are filtered out, but in contrast, no more additional frames are placed elsewhere. %
Especially due to the often quadratic run time complexity of global methods such as COLMAP, the reduction of a few redundant images can make a significant difference with large data sets. %
For a sequence with initially 1,056 images, our approach leads to a reduction in images of \approxi 5\percent especially in regions with little camera movement. %
This reduces the processing time from 1,510.24\,min\xspace to 1,388.25\,min\xspace which corresponds to a decrease of 8.1\percent. %
For this, we measured the individual duration of all steps in COLMAP's 3D reconstruction pipeline. %
Since the speedup is divided into 10.52\% for the sparse reconstruction and 4.95\% for the dense reconstruction, especially feature matching and sparse mapping seem to benefit from our filtering in advance, as the savings are disproportionately high. %
%sparse of: 955,105 min
%sparse 1fps: 1.055,625 min
%dense of: 433,146 min
%dense 1fps: 454,618 min
%of: 1.388,251 min
%1fps: 1.510,243 min
\begin{figure}[h!]
    \small ~~~~~~$I_{1}$~~~~~~~~~~~~~~~~~~~~~~~~$I_{2}$~~~~~~~~~~~~~~~~~~~~~~~~$I_{3}$~~~~~~~~~~~~~~~~~~~~~~~~$I_{4}$~~~~~~~~~~~~~~~~~~~~~~~~$I_{5}$\\
    \includegraphics[width=0.19\textwidth]{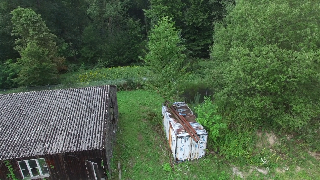}
    \includegraphics[width=0.19\textwidth]{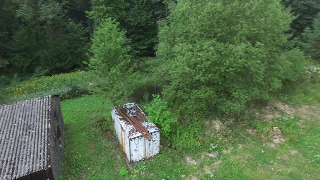}
    \includegraphics[width=0.19\textwidth]{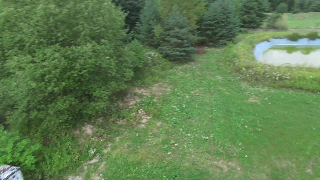}
    \includegraphics[width=0.19\textwidth]{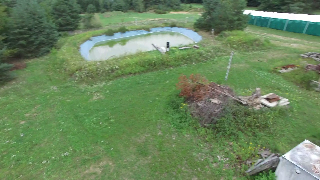}
    \includegraphics[width=0.19\textwidth]{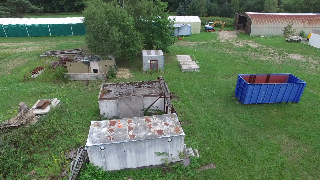}\\
    \vspace{-2em}
    \centering
    \includegraphics[width=\textwidth]{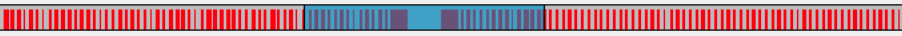}\\
    \caption{As can be seen in the images $I_{1}$ to $I_{5}$, the camera rotates quickly within \approxi 2\s by 180 degrees and therefore causes significant motion blur. % 
    The timeline below shows the entire sequence, with each red bar corresponding to an image that is considered sharp. %
    The blurred images shown above are located in the middle of the highlighted area and are successfully detected and excluded. %
    }
    \label{fig:example_blur}
\end{figure}

To further avoid processing images that do not provide novel content, we have developed a plugin that can detect and replace images affected by motion blur. %
The sequence we use for this shows usable images for most of the flight, but it includes a section where the camera rotates 180 degrees in about 2\s and the corresponding images become unusable due high degrees of motion blur. %
In \Cref{fig:example_blur} the corresponding section is shown, once by exemplary images and on the timeline below as an area marked in blue. %
Each red bar corresponds to an image that is considered sharp enough. As can be seen here, the area with motion blur is reliably detected and no images are selected here. %
However, it is also visible that key frames accumulate around this point. %
This occurs because although our plugin searches for a spatially close replacements, the search for better neighbouring images on this explicit sequence does not work optimally due to the fact that the blurred section is too long and therefore no better images exist. %
When selecting replacement images, the current image density should be taken into account in the future or the replacement should be made optional.%
%It may also be advisable to only exclude blurred images without searching for replacements, as this already works and reliably prevents the processing of such images. %

\subsection{Semantic segmentation for masking potentially moving objects}  
\begin{figure}[h!]
    \small  ~~Input image ~~~~~~~~~~Semantic segmentation ~~~~~~~~~ Binary mask 
    \centering
    \includegraphics[width=0.25\textwidth,height=0.14\textwidth]{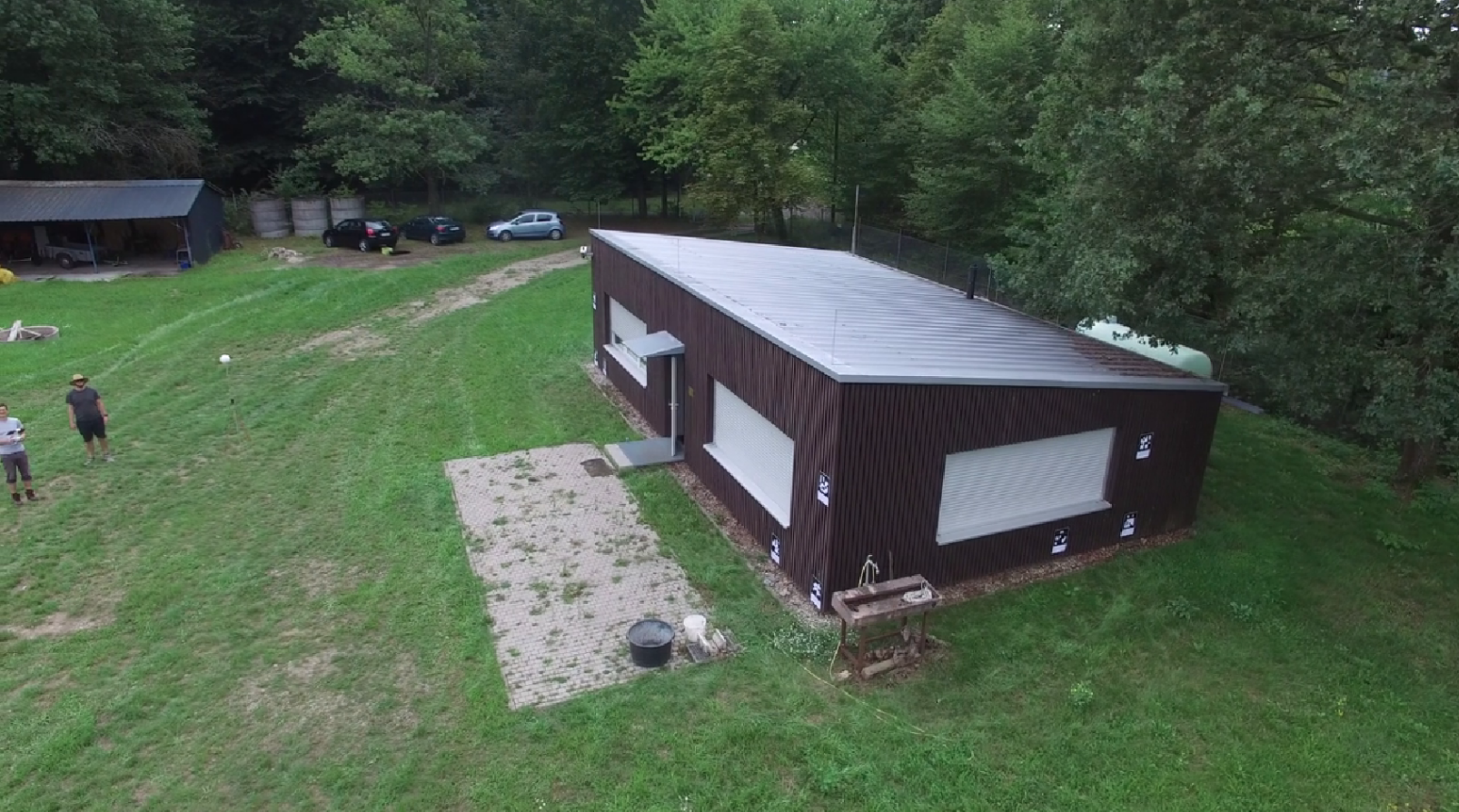}%
    \includegraphics[width=0.25\textwidth,height=0.14\textwidth]{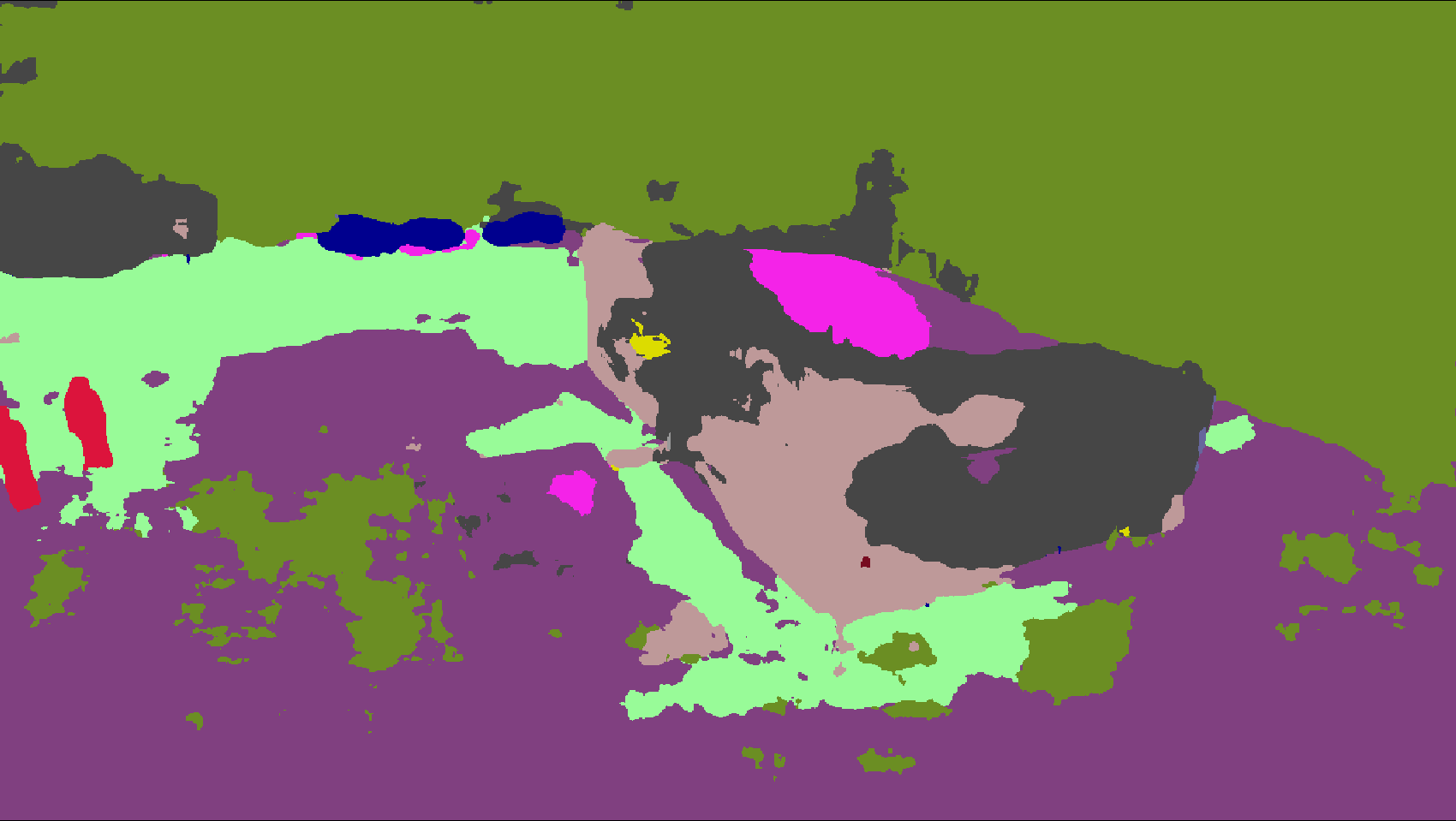}%
    \setlength{\fboxsep}{-0.5pt}%
    \setlength{\fboxrule}{0.5pt}%
    \fbox{\includegraphics[width=0.25\textwidth,height=0.14\textwidth]{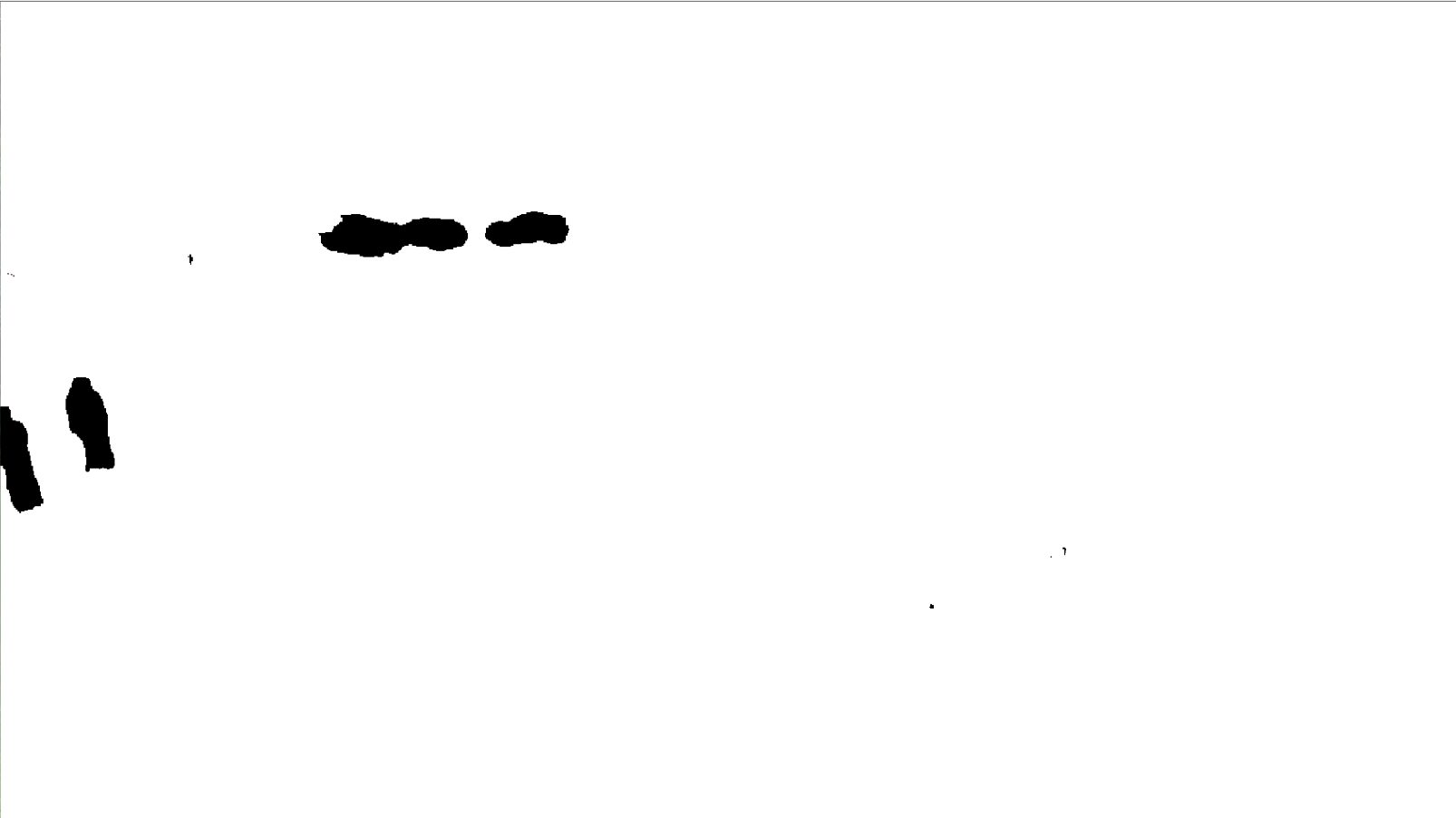}}%
    \vspace{-1mm}
    \includegraphics[width=0.25\textwidth,height=0.14\textwidth]{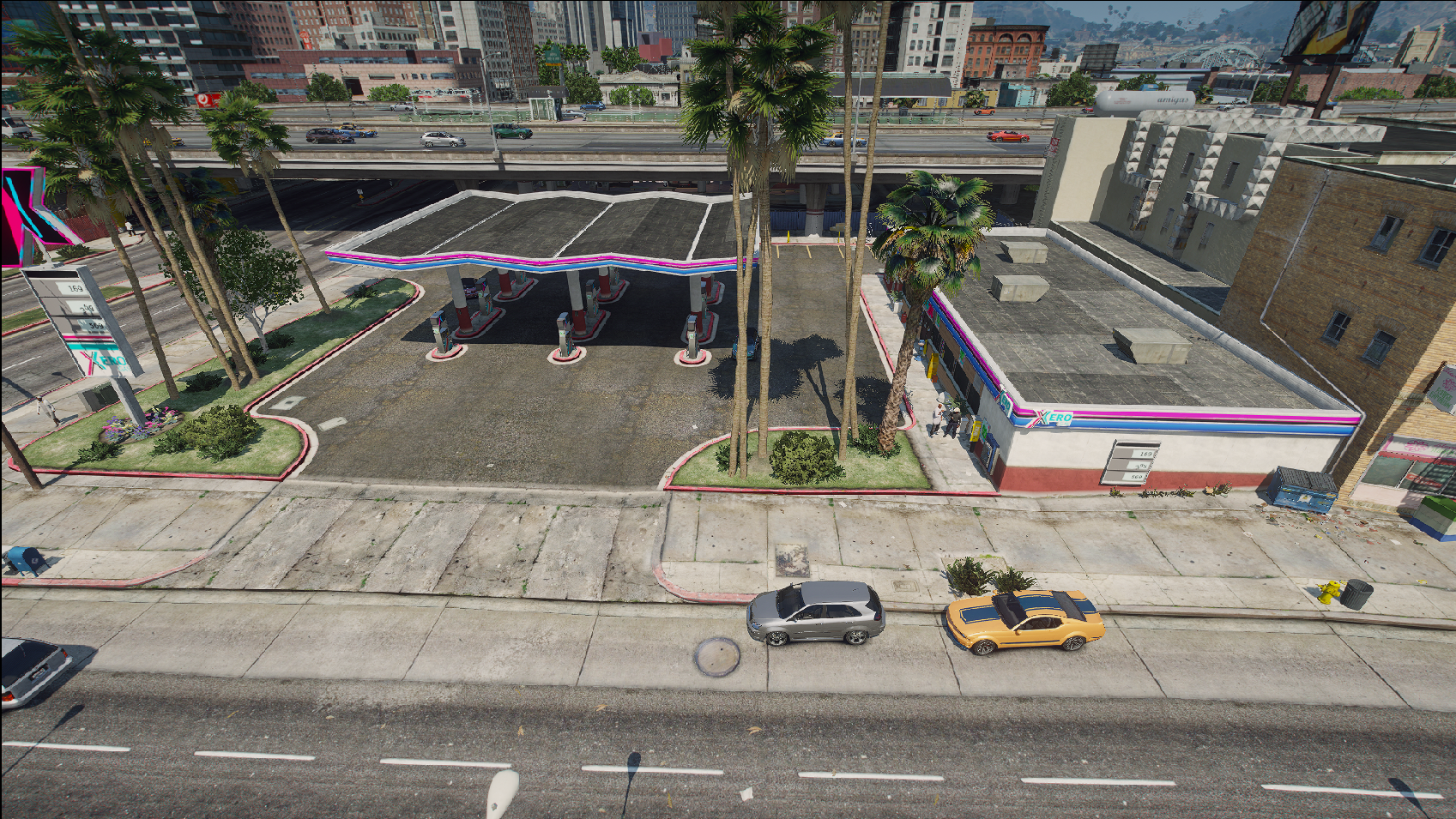}%
    \includegraphics[width=0.25\textwidth,height=0.14\textwidth]{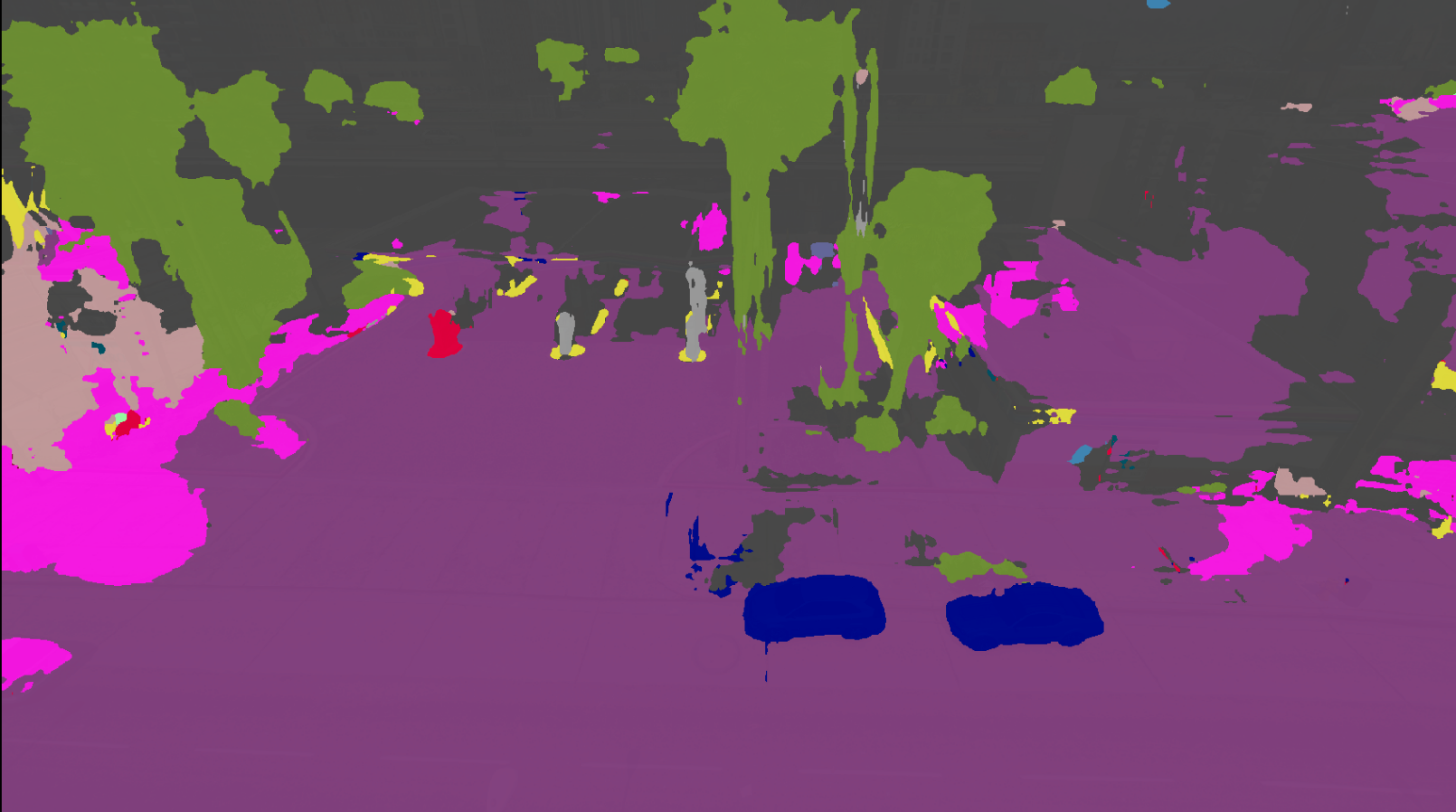}%
    \setlength{\fboxsep}{-0.5pt}%
    \setlength{\fboxrule}{0.5pt}%
    \fbox{\includegraphics[width=0.25\textwidth,height=0.14\textwidth]{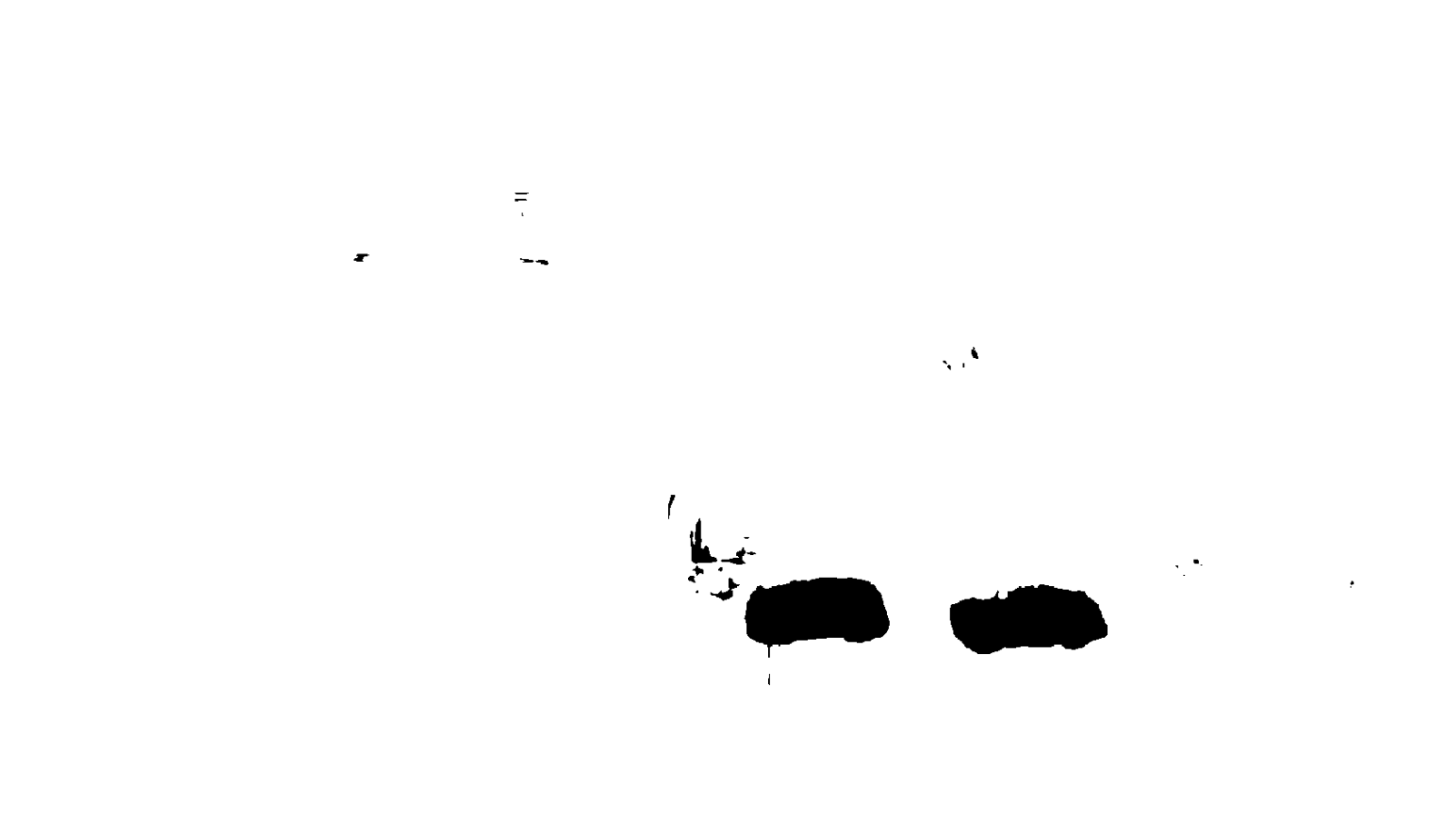}}%
    \caption{Examples of our semantic segmentation plugin. %
    From left to right: input image, semantic segmentation and binary mask excluding the classes car and person. %
    }
    \label{fig:semantic_examples}
\end{figure}

In addition to the reduction to the most significant images, our framework has as a second type of plugins for the enrichment with additional information. % 
Currently, we focus on masking challenging image areas through semantic segmentation.  % 
For this purpose, we provide a plugin that allows deep learning models with different depths and architectures to be imported and applied to the images. %
Since there are currently hardly any data sets for semantic segmentation of aerial images from oblique view, all neural networks for this purpose were trained on the Cityscapes data set. %
Being an autonomous driving data set, the domain gap with regard to the camera angle is in some cases large, which can also be seen in the results. %
However, as visible in \Cref{fig:semantic_examples}, reasonable results are achieved for relatively flat camera angles. %
Vegetation is reliably detected from all camera angles, but due to only one class available, a distinction between not problematic vegetation and trees moving in the wind is not possible. %
Since we filter out all objects that could be moving, objects such as parked cars are also excluded, which may be undesirable depending on the type of application. %

In our experiments with aerial images, masking people and vehicles did not lead to a quantitatively better quality of the 3D models. %
We attribute this to sub-optimal segmentations, as the neural networks we use were not trained for oblique-view aerial imagery and thus contain a high rate of errors when the camera angle is very different from that of the training data. %
In addition, COLMAP already has photometric and geometric consistency verification techniques that can compensate for some artefacts induced by movement \cite{schoenberger2016mvs}. %
During our tests, however, we noticed that especially scenarios with a large amount of movement, e.g. next to a motorway, often did not yield a useful sparse reconstruction. %
Although in some cases a useful sparse reconstruction can be obtained by starting the reconstruction several times, it is here in particular that we see potential for semantic segmentation to ease 3D reconstruction. %
Once the sparse reconstruction succeeds, the results for the dense reconstruction are quite good, as can be seen from sequence C in \Cref{tab:quantitative_results}. %
However, the drop in $\delta_{0.05}$ accuracy of the depth maps compared to both sequences A and B with less movement within the scene is worth noting. %
This is probably because moving objects are filtered out in the reconstruction but are present in the ground truth depth maps. %
\section{Conclusion \& Future Work}
\label{sec:conclusion}

In this paper, we presented a new framework for pre-processing image sequences to facilitate 3D reconstruction, which we publish open source using the MIT licence. %
The modular architecture allows for easy customisation and integration of already existing algorithms by encapsulating them in plugins using our simple interfaces. %
This provides the opportunity to participate in the development of the core application or individual plugins. %
So far, we have developed plugins that can detect redundant frames, replace blurred images and also mask potentially moving objects. %
As shown in the previous section, these baseline methods are already capable of accelerating a downstream 3D reconstruction by 8.1\percent through reducing redundant images. %
By creating a dedicated workflow for batch processing, large quantities of images can be edited with little effort, e.g. by specifying frame rate, resolution, cropping and the utilised plugins only once and then processing multiple videos with them.  %
In the future, we focus on extending our current plugins and the development of further ones. %
One possibility here would be to perform a real-time \gls*{SLAM} algorithm such as ORB-SLAM3 \cite{9440682} and extract the key frames or incorporate metadata present in the input videos. %
Similarly, we plan to further enhance the semantic segmentation plugin by incorporating additional models trained on other data sets, as it has been shown that the domain gap can be very high for aerial imagery depending on the camera angle. %
We are convinced that the framework we provide offers great added value, especially when more plugins become available in the future. %

%
%
%


\bibliographystyle{splncs04}
\bibliography{bibliography}

\end{document}